\documentclass[10pt,twocolumn,letterpaper]{article}

\usepackage{iccv}
\usepackage{times}
\usepackage{epsfig}
\usepackage{graphicx}
\usepackage{amsmath}
\usepackage{amssymb}

\usepackage{tabls}
\usepackage{stfloats}
\usepackage{tabularx}
\usepackage{booktabs}
\usepackage{xcolor}
\usepackage{amsmath, bm, subfigure, epstopdf, url, pifont, overpic, cases}
\usepackage{latexsym, amssymb, bbding, multirow, makecell,  diagbox, enumitem, caption}
\usepackage{array, enumitem, soul, algorithm, algpseudocode}

\usepackage{arydshln}
\usepackage{multirow, multicol}
\usepackage{adjustbox}

\newlength \g

\usepackage[pagebackref=true,breaklinks=true,letterpaper=true,colorlinks,bookmarks=false]{hyperref}

\iccvfinalcopy %

\def\iccvPaperID{10} %

\begin{document}

\title{Transformer-Based UNet with Multi-Headed Cross-Attention Skip Connections to Eliminate Artifacts in Scanned Documents  }

\author{
  David Kreuzer  \\
  University of Applied Science \\
  Ulm\\
  \texttt{david.kreuzer@thu.de} \\ \\
  \and
  Michael Munz \\
   University of Applied Science \\
  Ulm\\
  \texttt{michael.munz@thu.de} \\
}
\maketitle

\begin{abstract}
The extraction of text in high quality is essential for text-based document analysis tasks like Document Classification or Named Entity Recognition. Unfortunately, this is not always ensured, as poor scan quality and the resulting artifacts lead to errors in the Optical Character Recognition (OCR) process. Current approaches using Convolutional Neural Networks show promising results for background removal tasks but fail correcting artifacts like pixelation or compression errors. For general images, Transformer backbones are getting integrated more frequently in well-known neural network structures for denoising tasks. In this work, a modified UNet structure using a Swin Transformer backbone is presented to remove typical artifacts in scanned documents. Multi-headed cross-attention skip connections are used to more selectively learn features in respective levels of abstraction. 
The performance of this approach is examined regarding compression errors, pixelation and random noise. An improvement in text extraction quality with a reduced error rate of up to 53.9\% on the synthetic data is archived. The pretrained base-model can be easily adopted to new artifacts. The cross-attention skip connections allow to integrate textual information extracted from the encoder or in form of commands to more selectively control the models outcome. The latter is shown by means of an example application.

\end{abstract}

\section{Introduction}
As the importance of Natural Language Processing (NLP) and  Natural Language Understanding (NLU) increases and language models are getting more powerful \cite{10.48550/arxiv.2111.01243}, a wide range of possibilities are created to optimize outdated processes. Since many processes cannot yet be completely handled digitally, this oftentimes still involves working with scanned documents, prints and faxes. To extract textual information in these cases, the need of high quality OCR results is not always fulfilled, leading to bad text quality and forfeiting machine learning model performance due to input errors \cite{Maekawa19}. As the errors in scanned artifacts can arise from a variety of different artifacts, improving image quality is not trivial. For that reason, this problem is frequently tackled by correcting errors in the text after the extraction \cite{bart20, Sengupta21, Zhang20} instead of enhancing the document quality. Unfortunately, these methods fail correcting new entities like infrequent names, addresses or contact data (phone, fax, email addresses), which is critical for information extraction tasks.  
Nevertheless, there are approaches to improve text quality in documents using Generative Adversarial Networks (GANs) or Autoencoders based on Convolutional Layers  \cite{Maekawa19,10.1109/icpr.2018.8546199, 10.1109/icirca54612.2022.9985695, 10.1109/icrito56286.2022.9964743}. However, these works mainly focus on the removal of noisy background artifacts, using a relatively small dataset with low data variance. In practice, these errors are only a fraction of the problem. Other than noise, the OCR result often suffers from compression errors, image down-sampling, and fading characters.
Compression errors and super resolution tasks are deeply studied for images \cite{10.48550/arxiv.2108.10257, 10.48550/arxiv.2205.04437, conde2022swin2sr}. Yet, very little consideration has been given to these new methods regarding scanned documents.
Vision Transformer models show great potential achieving state of the art results \cite{10.48550/arxiv.2101.01169}, as can be observed in similar domains. Especially Swin Transformer models show good results when dealing with textual information in images \cite{DBLP:journals/corr/abs-2103-14030, 10.48550/arxiv.2111.15664}. 
In 2022 Fan et. al \cite{10.48550/arxiv.2202.14009} published good results using a Swin Transformer based UNet (SUNet) to denoise images. The network structure was inspired by Cao et. al \cite{10.48550/arxiv.2105.05537}, optimizing the up-sampling algorithm. UNet architectures are widely used in image processing applications and maintain a margin open for variation. While Fan et. al and Cao et. al focus on newly designed down-sampling and up-sampling modules based on a Swin Transformer, others examine the influence of varying skip connection types \cite{10.48550/arxiv.2109.04335}.      
Usually, feed forward layer are used to bridge information from Encoder to Decoder. However, feed forward layer have the disadvantage that they are not suitable for processing sequential inputs and lack in the ability to selectively learn relevant information. Especially in whole images being bridged, this does not seem to be the best option. Furthermore, due to the linguification of many machine learning applications, in-cooperating sequential data in form of speech can be a useful requirement for such models. 
To address these challenges, this paper proposes to replace feed forward skip connections of the UNet with cross-attention modules. The use of these modules to gain flexibility in input and output dimension has already been examined in a publication of Andrew Jaegle et al. \cite{10.48550/arxiv.2103.03206}. 
Since multimodality plays a role especially in document processing, these results support the decision to replace most skip connections. \newline
In this paper a modified UNet with Swin Transformer backbone to address the removal of common artifacts in scanned documents is presented. %
To evaluate the direct influence on Machine Learning tasks, the OCR improvement after cleaning the documents is evaluated. 
The contributions of this paper are listed as follows:
\begin{itemize}
    \item The use of a synthetic dataset to pre-train a Swin Transformer UNet for document enhancement.
    \item Presentation of modified cross-attention skip connections to improve the information flow.
    \item Evaluation of the model performance with regard to the improvement of the text extraction quality. 
    \item Visual investigation of the model adaptability by fine-tuning on a public dataset.
    \item The incorporation of textual information to control the task to be executed by exploiting the adapted skip connections.
\end{itemize}

\section{Materials and Methods} 

\subsection{Model Structure}
\begin{figure}[H]	
\centering
\includegraphics[width=0.4\textwidth]{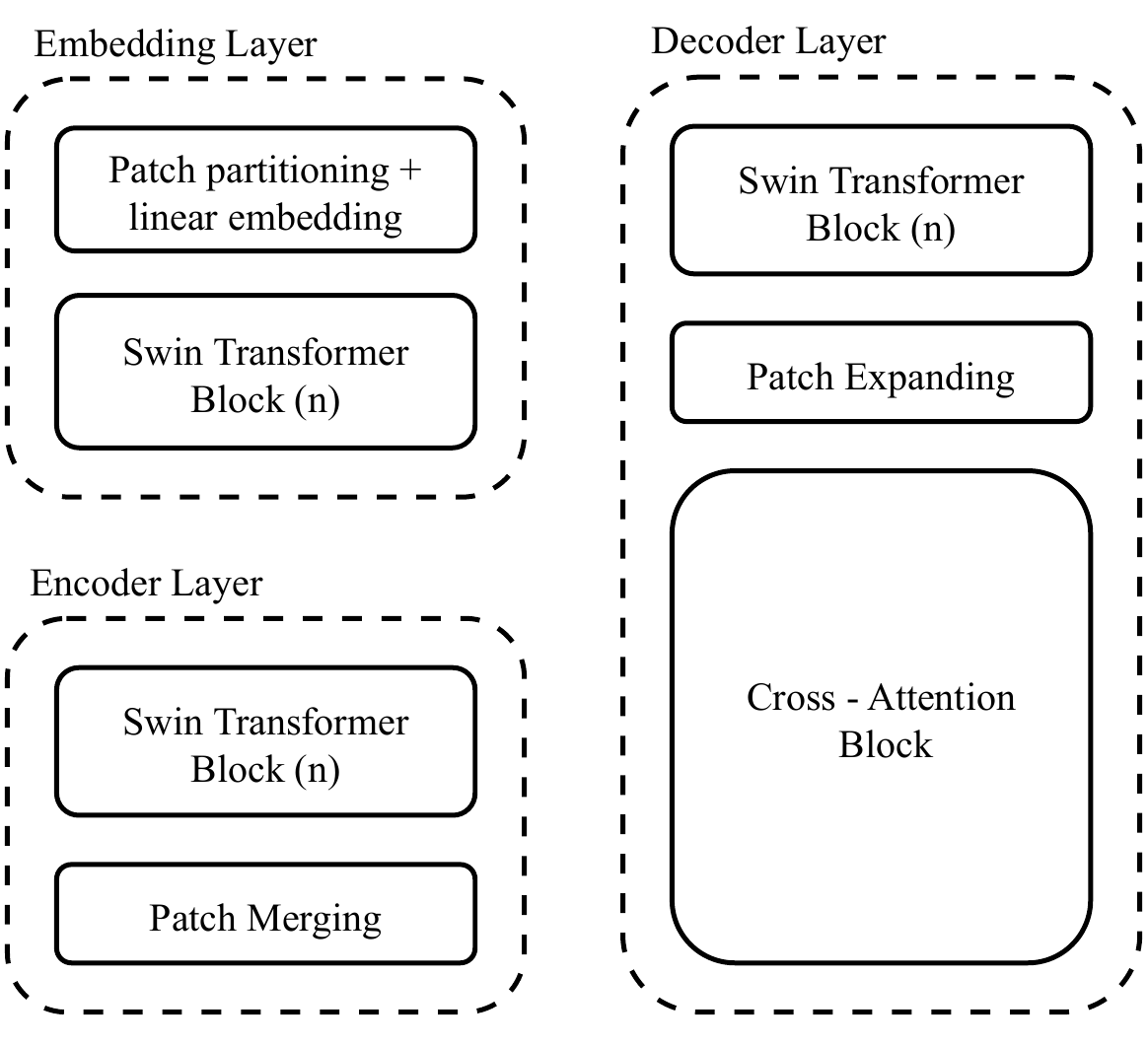}
\caption{Design of the basic encoder layer, decoder layer and embedding layer used in the model. In the embedding layer the patch partitioning and linear projection of patches to embedding vectors is done, followed by $n$ repeated Swin Transformer blocks.  The encoder layer consists of $n$ repeated Swin Transformer blocks followed by the patch merging module. From this module emerges the dimension reduction. The decoder block includes the cross-attention block followed by the Patch-Expansion (upsampling) module and $n$ repeated Swin Transformer blocks. The information flow in the encoder passes from top to bottom and in the decoder from bottom to top.}
\label{fig:encoderdecoder}
\end{figure}

The proposed model structure builds on the Swin Transformer UNet architecture presented in \cite{10.48550/arxiv.2105.05537}. 
The encoder consists of Swin Transformer blocks (STB) using multi-head self attention modules with regular (W-MSA) and shifted windowing (SW-MSA) \cite{10.1109/iccv48922.2021.00986}. The hierarchical patch-merging module substitutes convolutional operations of conventional UNet structures and provides the desired down-sampling effect. 
The structure of a decoder layer is shown in Fig. \ref{fig:encoderdecoder}. As in various pre-tests, the patch expanding method presented in \cite{10.48550/arxiv.2202.14009} showed better results than the approach used in \cite{10.48550/arxiv.2105.05537}, this method was used, leaving room for further investigation. In combination with a repeated STB, this module remains the central component of the decoder.  
In the last layer, a $4$x up-sampling block is used. Additionally, up-sampling is facilitated by bridging the original state in to the decoder. This is done by using multi-headed cross-attention skip connections.  

\subsubsection*{Skip Connections}
\label{subsubsec:skipconnections}
\begin{figure}[H]	
\centering
\includegraphics[width=0.4\textwidth]{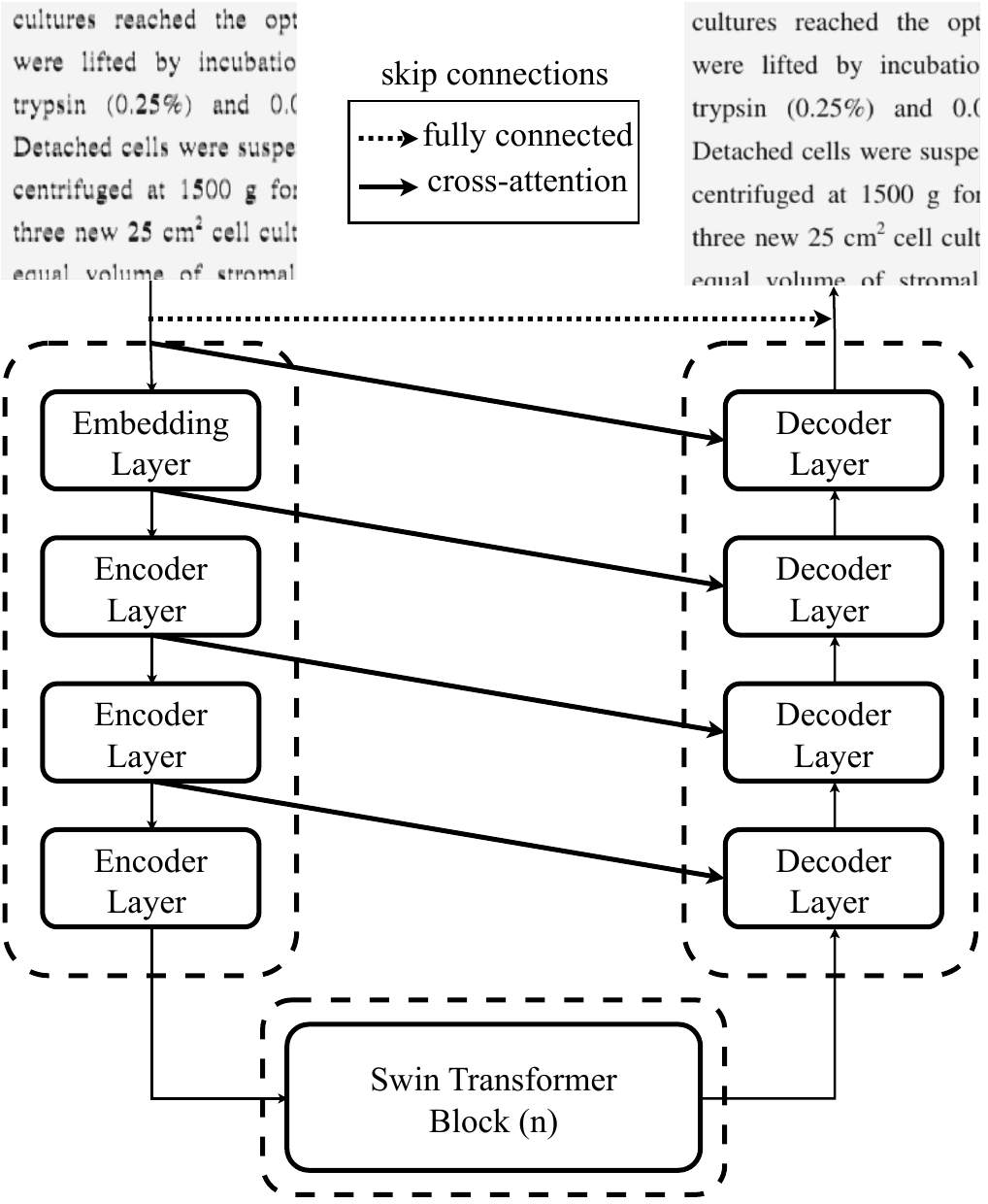}
\caption{Structure of the proposed model and its skip connections. The skip connections allow for the direct flow of information between non-adjacent layers in the network, improving gradient flow and enabling better training. The conventional skip connection is marked as dotted arrow while the cross-connections are represented by solid lines. An examplary pixelated input can be seen on the left side, being fed to the model. The output with improved quality is shown on the right side.}
\label{fig:networkskipconnections}
\end{figure}

Although conventional skip connections are, in most cases, represented as feed forward layer, merging concatenated information, multi-headed cross-attention modules are considered. 
This enables the model to selectively focus on relevant information, which is crucial in scanned documents. Especially artifacts like pixelations and compression errors occur concentrated close to the text, while unwritten parts of the document do not suffer from these errors. 
Another novelty is, that the information of the encoder is bridged to deeper decoder states, see Fig. \ref{fig:networkskipconnections}. Ideally, this gives access to parts of the image that can be adopted from pixels and features extracted from the encoder. 

\subsubsection*{Cross-Attention}
The idea of mapping a latent array to another latent array using cross-attention has already been utilized in the Perceiver architecture \cite{10.48550/arxiv.2103.03206}.
In the multi-headed cross-attention process key and value, defined as $K = S_eW^K$ and $V = S_eW^V$, are linear projections of the encoder state $S_e$ which represents the \textit{context} being fed to the module. It allows to choose context matrices of high sizes, since the \textit{latent} dimension of the representation, being passed in by the previous decoder layer, is maintained. Here, the query vector $Q = S_dW^Q$ is a linear projection of the current decoder state $S_d$. In the model, eight dimension heads $h$ and a head dimension of $64$ are used resulting in an inner dimension $d_{inner} = 512$ and the learnable parameter $W$, where $W^K, W^V \in \mathbb{R}^{d_{S_e} \times d_{inner}}$. 
The attention between $Q_iK_iV_i$ is defined by \cite{10.48550/arxiv.1706.03762}, where $i$ indicates the index of the current attention head. For one head applies
\begin{equation}
    A(Q_i,K_i,V_i) = \mathrm{softmax}\Bigg(\frac{Q_iK_i^T}{\sqrt{d_k}}\Bigg)V_i,
\end{equation}
where $d_k = \frac{d_{S_e}}{h}$ is the dimension of the encoder state. 
It should be emphasized that $K$ and $V$ can be enriched with additional information without changing the latent dimension. 
\begin{figure}[h!]	
\centering
\includegraphics[width=0.5\textwidth]{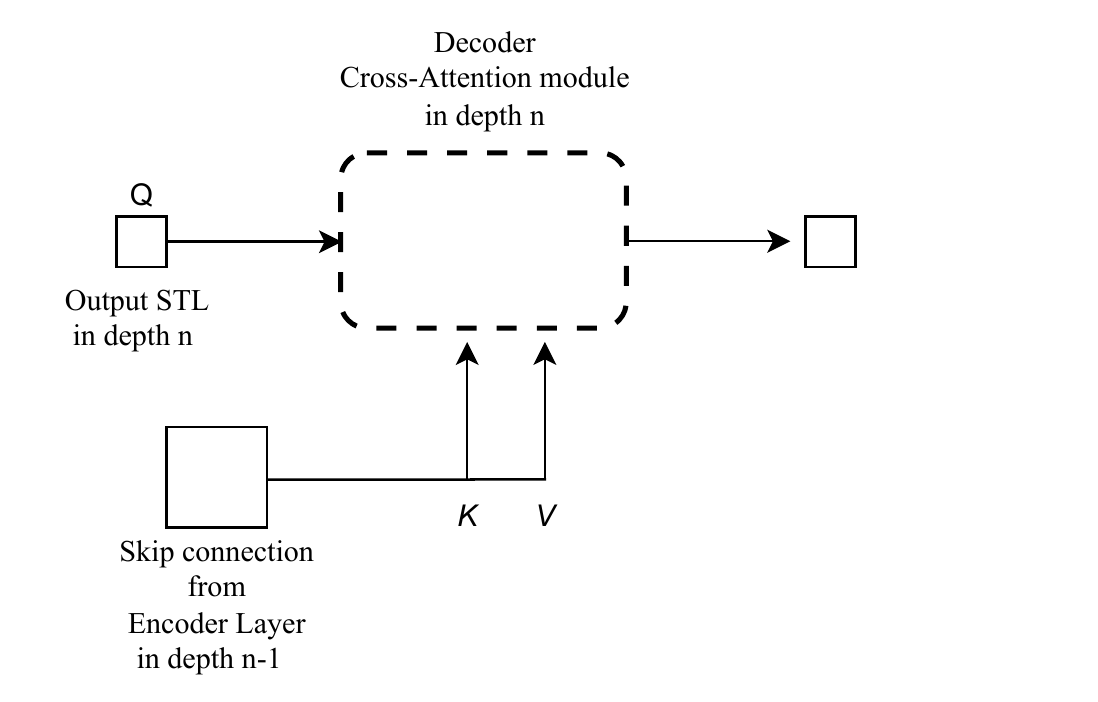}
\caption{Cross-attention module of the decoder block in depth n: The Output of the Swin Transformer Layer (STL) in the same block  which represents the latent array of the cross-attention module is decisive for the modules output dimension. The context is provided by features from the encoder in depth n-1.}
\label{fig:crossattention}
\end{figure}

\subsection{Dataset}
Training the model requires both, documents with artifacts and their clean counterparts. Unfortunately, many big datasets contain only one of the two. 
In previously published works, a small dataset is often used. Souibgui and Kessentini presented a DE-GAN to tackle the problem in 2020 \cite{DBLP:journals/corr/abs-2010-08764}. However, the dataset used also lacks artifact variety. Furthermore, the model, like most transformer models, requires a high amount of training data to realize its full potential. For that reason, a synthetic dataset is created, allowing pre-training the model to evaluate each artifact separately. Clean ground truth documents are obtained from the PubLayNet dataset \cite{DBLP:journals/corr/abs-1908-07836}, which is a big dataset for document layout analysis tasks. PubLayNet was chosen because it includes multiple document structures, fonts and font sizes, which is an important prerequisite to not overfit on the dataset. Artifacts, that are synthetically added are compression errors and pixelation errors in words and noise. To save resources, the images are cropped to a height and width of 256 pixels to study the performance of the model in smaller image cutouts.
\begin{figure}[H]	
\centering
\includegraphics[width=0.335\textwidth]{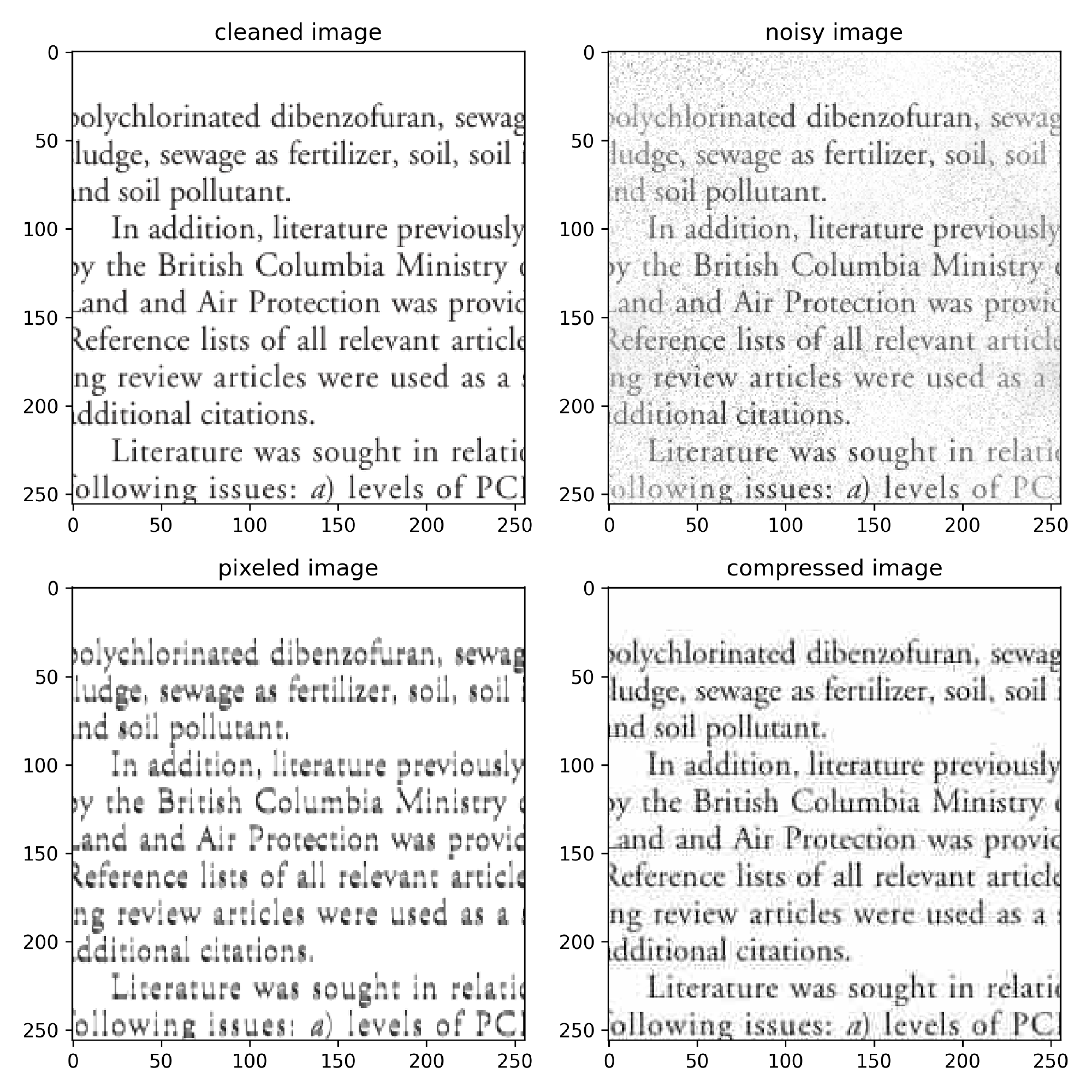}
\caption{Artifacts used for training: An example of a real image can be seen on the top left, pixelation errors are presented in the bottom left, an example of a noisy image is shown on the top right and compression errors are shown on the bottom right.}
\label{fig:artifacts}
\end{figure}
\subsubsection*{Compression Errors and Pixelation}
In comparison to noise artifacts, pixelation and compression errors are found locally in the vicinity of text. For that reason, those artifacts are more common for scanned documents when comparing it to other image processing tasks.

Especially similar characters can be more easily confused during the OCR by the influence of these artifacts, as can be seen in Fig. \ref{fig:artifacts}. An obvious problem is that the character \textbf{i} is being confused with \textbf{l}, since the space between the point and the stem is easily merged by artifacts of the size of only a few pixels. The framework to generate the synthetic training data tackles exactly this problem. Besides common compression errors, morphological operations are used to simulate these errors in specific. In particular, gray value dilation generates the desired effects.    
\subsubsection*{Noise}
In contrast to the previously regarded artifacts, noise applies to the whole document. This entails the risk of whole words or letters being covered. 
These types of artifacts often arise in the process of scanning physical documents. It can either be the case that the documents are placed incorrectly on the scanner, the scanner has incorrect settings, or the documents have been soiled.
To add noise to the documents, the imgaug library\footnotemark is used. In 10 percent of the documents, distortion effects are integrated by applying pairwise affine transformations to the original images. The noise is generated by using a combination of cloud noise and speckle noise. 
{\footnotetext{https://github.com/aleju/imgaug}

\subsection{Training}
The model is trained for 20 epochs on 8 NVIDIA RTX A6000 graphic cards. The AdamW optimizer is used with a decay of 0.1.  Additionally, a linear learning rate scheduler modifies the learning rate during an epoch in the interval of $1\mathrm{e}^{-5}$ to $2\mathrm{e}^{-7}$. A batch size of 8 samples per GPU can be afforded. The training set consists of 167900 samples and the whole training process runs for 2 days on the given setup.
The Swin Transformer encoder uses a patch size, as well as a window size of 4 and an embedding dimension of 64. The depth is chosen to be 4. 
The images are scaled to a range between $(0,1)$ for training. Each artifact is treated separately for better evaluation. The images are cropped to a sizes of $(256, 256, 3)$. The image sections are boxes with diagonal spanned by $(x_1, y_1)$ and $(x_2, y_2)$, with chosen values: $x_1$\,$=$\,$y_1$\,$=$\,$0$~; $x_2$\,$=$\,$y_2$\,$=$\,$256$ and $x_1$\,$=$\,$y_1$\,$=$\,$256$~; $x_2$\,$=$\,$y_2$\,$=$\,$512$ to represent margins and middle parts of the document.      

\subsubsection*{Loss}
The loss function for training is a combination of the L1 pixel-distance and the perceptual loss. Combining both loss functions brings the advantage, that also extracted high level features are integrated in the training process. Especially for image reconstruction and super resolution tasks, the perceptual loss can bring benefits and is already used for similar tasks \cite{DBLP:journals/corr/JohnsonAL16, 10.48550/arxiv.2108.10257}. This leads to the following loss function:
\begin{equation}
    L =  \underbrace{\frac{1}{n} \sum_{i=1}^{n} |x_i - \hat{x}_i|}_{\text{L1 Loss}} + c \underbrace{\sum_{j} w_j \left\| \phi(x)_j - \phi(\hat{x})_j \right\|^2}_{\text{Perceptual Loss}},
\end{equation}
where $n$ is the number of pixels, $x_i$ and $\hat{x}_i$ are the pixel values at the pixel position $i$, respectively. Furthermore $\phi(x)$ is the function extracting features. The individual feature levels are assigned weights using the parameter $w_j$. The perceptual loss is factored to a smaller extent into the total loss by using the constant $c$, which is chosen $0.1$.  
 
\subsection{Experiments}
In the paper the capability of the model, dealing with the presented artifacts is evaluated. The artifacts are generated synthetically, which opens the opportunity to focus on each artifact separately. Especially the influence on the text extraction is considered. 
\subsubsection*{Improvement of Text Extraction}
To calculate the improvement, the OCR result of the \textit{clean} document is considered as ground truth, meaning that errors arising from the OCR algorithm itself processing the document without artifacts are ignored. On clean documents, however, state of the art OCR algorithms work very reliably. The error is negligible as only the difference between the text extraction before and after cleaning the document is of interest.
This leads to the adapted word error rate.
\begin{equation}
    AWER = \frac{\text{\# word matches }}{\text{\# total extracted words (ref. image)}},
\end{equation}
where the number of matching words is calculated by comparing the extracted text before and after cleaning the document based on the words position.  With this metric, the improvement in extraction quality can be analyzed directly. Nevertheless, attention must be drawn to the fact that especially for cropped images,  errors emerging from truncated word parts can occur in the ground truth, which cannot be corrected reliably. Especially OCR methods that are based on a vocabulary tend to make mistakes in that case. 
\newline
For some applications images are filtered with standard image processing methods, like noise reduction filters, before extracting the text. For that reason the improvement of the extraction quality is compared to a filtered image, preprocessed by the non local means (NLM) filter. Among standard filters for noise reduction, the non-local means filter has the advantage that it excels in preserving fine details and structures in the image while reducing the noise \cite{10.1109/cvpr.2005.38}. Also its a robust method which is used broadly in many domains, as for example for cleaning medical images \cite{10.1155/2012/438617}.    
\subsubsection*{Adaptability}
As the model is trained on synthetic data, the performance on other datasets is of interest. To investigate the adaptability of the model, a known dataset containing noisy scanned documents is used. The \textit{scanned noisy documents} dataset includes a different type of artifacts and also consists of 55 simulated and 73 real noisy scanned documents. The documents are again re-sampled and cropped from $1550 \times 555$ to a size of $256 \times 256$ pixels. The effort to fine-tune the model for the dataset is evaluated. To prove, that due to the implemented cross-attention skip connections additional information can easily be integrated, the model is fine-tuned on opposing, noncumulative tasks whose execution is controlled by textual information given to the network additionally.     

\subsubsection*{Influence of Skip Connections}
The influence of the cross-attention skip connections is investigated in section \ref{res:cross-att}. To do so the model is compared to the SUNet presented in \cite{10.48550/arxiv.2202.14009}, which exclusively uses conventional skip connections. Both models are trained with the same configuration for the Encoder and Decoder. The training process is also unified. Besides the result, the convergence behavior is analyzed during training and fine-tuning. 


\section{Results}
\subsection{Visual Observations}
\begin{table*}[!htp]
\centering
\begin{tabular}{|c | c | c | c | c | c |} 
\hline
 \multirow{2}{*}{Artifact} & \multicolumn{3}{c|}{AWER (averaged)} & \multicolumn{2}
{c |}{Improvement}   \\ 
 \cline{2-6} 

  & original (\%)  &  non-local Means (\%) & our model (\%) & non-local Means (\%) & our model (\%) \\ [0.3ex]
 \hline
 Pixelation & 29.1 & 29.0 & \textbf{18.3} & 0.3 & \textbf{37.1}  \\ 

 Compression & 23.6 & 20.8 & \textbf{15.1} & 11.9 & \textbf{36.0} \\
 
 Noise & 26.9 & 22.8 & \textbf{12.4} & 15.2 & \textbf{53.9} \\
 \hline
\end{tabular}
\caption{$\overline{AWER}$ before and after cleaning the document as well as its improvement in percent. The improvement is giving the percentage by which the error is reduced applying the respective method. 
The first column shows the regarded artifact, the second column shows the absolute mean error made by the OCR engine of the scanned documents before enhancing them using the model. The third column gives the absolute mean errors made by the OCR after enhancing the image with the model and the improvement is calculated by calculating the relative reduction of errors using our document enhancing strategy. The highest impact can be observed cleaning noisy images.}
\label{tab:resultsartifacts}
\end{table*}
Fig. \ref{fig:prediction} ($1^{st}$ row) shows the effect on documents with typical artifacts that can occur when using low quality scan devices or old ink in a printer. As can be seen in the exemplary document excerpts, the model is able to restore the uniformity of the writing. Especially, unwanted pixel openings and closures are reduced by the network. Note, that the training data contains a variety of different fonts, which prevents overfitting on a specific one. 
The task in Fig. \ref{fig:prediction} (last row) is quite similar, as compression errors often lead to pixel closures in lower frequencies. Sometimes OCR engines fail due to similar effects. The average adapted word error rate $(\overline{AWER})$ when dealing with these artifacts can be seen in Tab. \ref{tab:resultsartifacts}. 
In the middle row, results regarding noisy images can be observed. When words are overlaid by the noise, the information cannot yet be recovered. Noise that stands out from the text is however properly removed.

\begin{figure}[htbp]
\centering
\begin{minipage}[b]{0.5\textwidth}
  \centering
  \includegraphics[width=\textwidth]{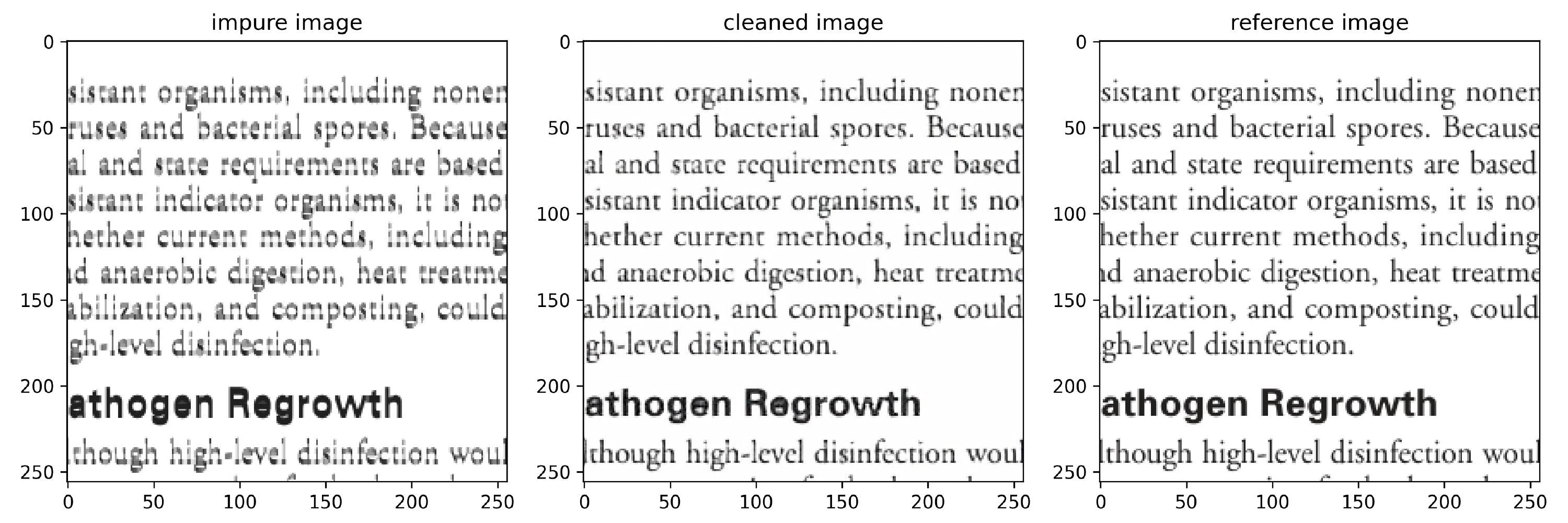}
\end{minipage}
\hfill
\begin{minipage}[b]{0.5\textwidth}
  \centering
  \includegraphics[width=\textwidth]{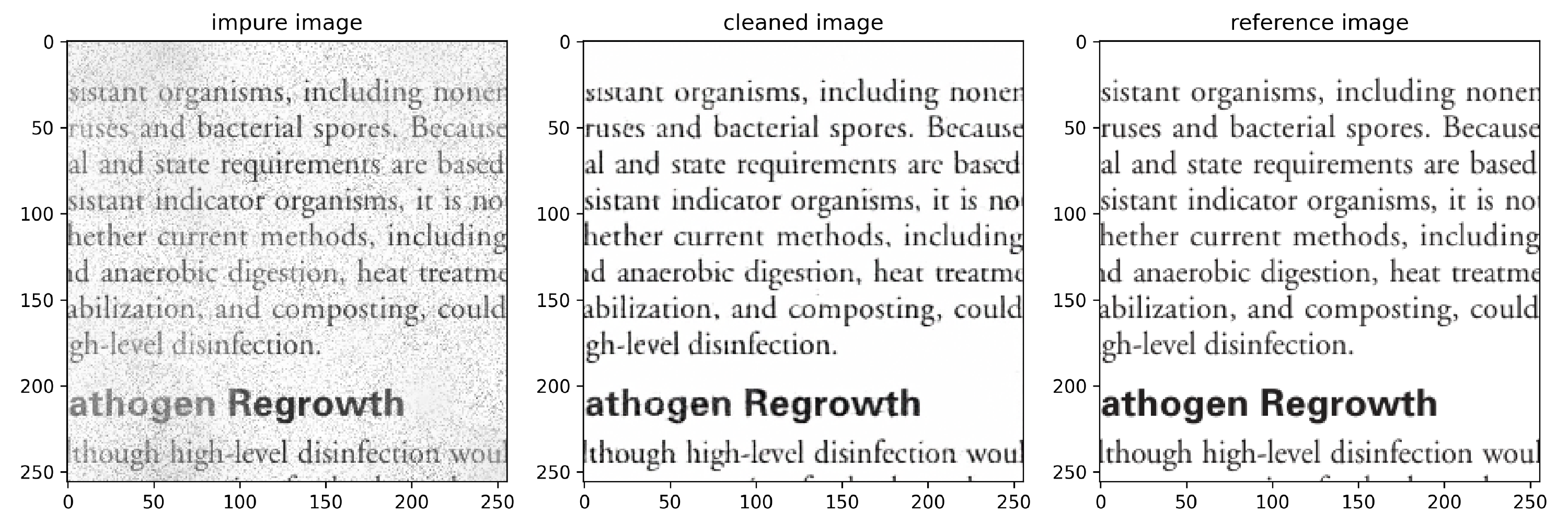}
\end{minipage}
\hfill
\begin{minipage}[b]{0.5\textwidth}
  \centering
  \includegraphics[width=\textwidth]{{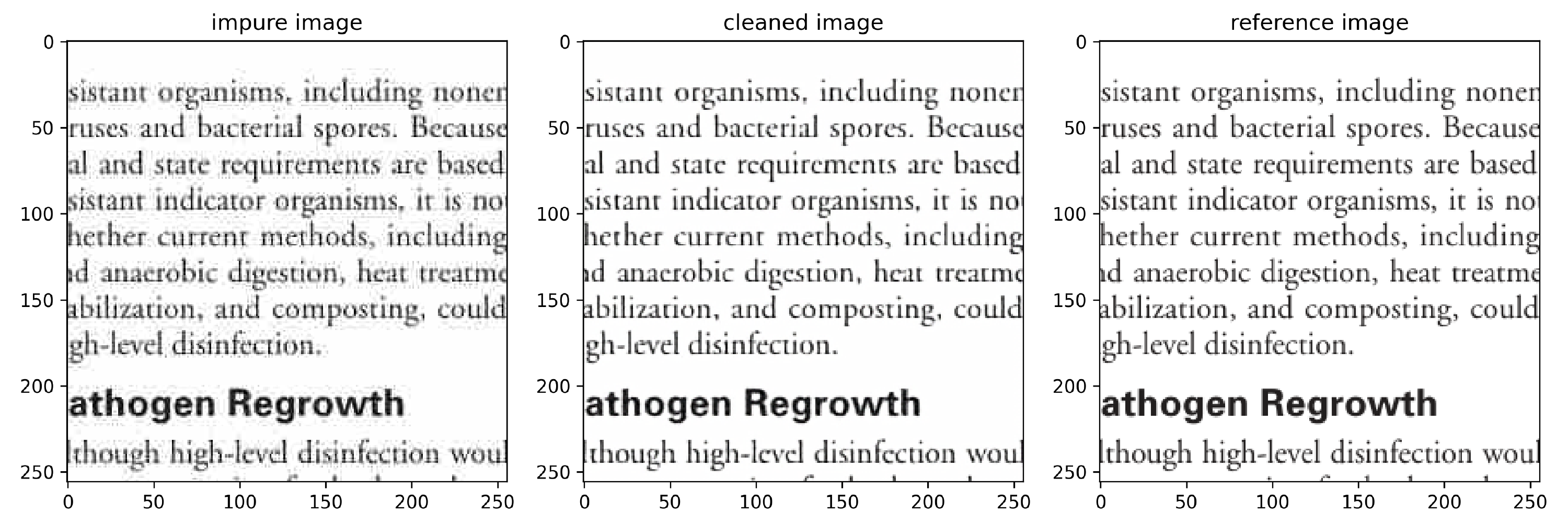}}
\end{minipage}
\caption{Visual results of the model inference on the given artifacts. In the first row the result of applying the cleaning process on pixelation errors. In the second row a noisy example is shown and in the bottom row the performance on reducing compression errors can be seen. In the first column the model output can be seen, which is the cleaned image. The third column presents the reference image that is used for training. In the middle column, the impure image with the respective artifact is presented.}
\label{fig:prediction}
\end{figure}
\subsection{Text Extraction}
As it is shown in Tab. \ref{tab:resultsartifacts} these types of artifacts have a big impact on the OCR result, as the whole document is synthetically modified and each word comprises structural changes. This leads to mean word error rates of up to 29.1\% $\overline{AWER}$, when dealing with pixelation errors. With the presented model, on average more than 37\% of these errors can be corrected. 
Errors caused by compression phenomena can be improved by 36\% of the overall words.  The best performance is achieved dealing with noisy images, correcting more than half ($53.9\%$) of the occurring errors.      

\subsection{Adaptability}
\label{res:fewshot}
To get great perfomance on a completely new dataset the base model has to be fine-tuned, since only a subset of common artifacts are covered by the synthetic dataset and it still needs to be extended. Nevertheless the pretrained model can be adapted easily to new tasks. To show the ability to adapt to new artifacts a dataset used in \cite{10.1109/icrito56286.2022.9964743} is utilized. 
Only 56 images with given artifacts are used for retraining the model for 2 epochs.
The fine-tuning process is accomplished in less than one minute on a single NVIDIA 3090Ti.

\begin{figure}[h!]	
\centering
\includegraphics[width=0.335\textwidth]{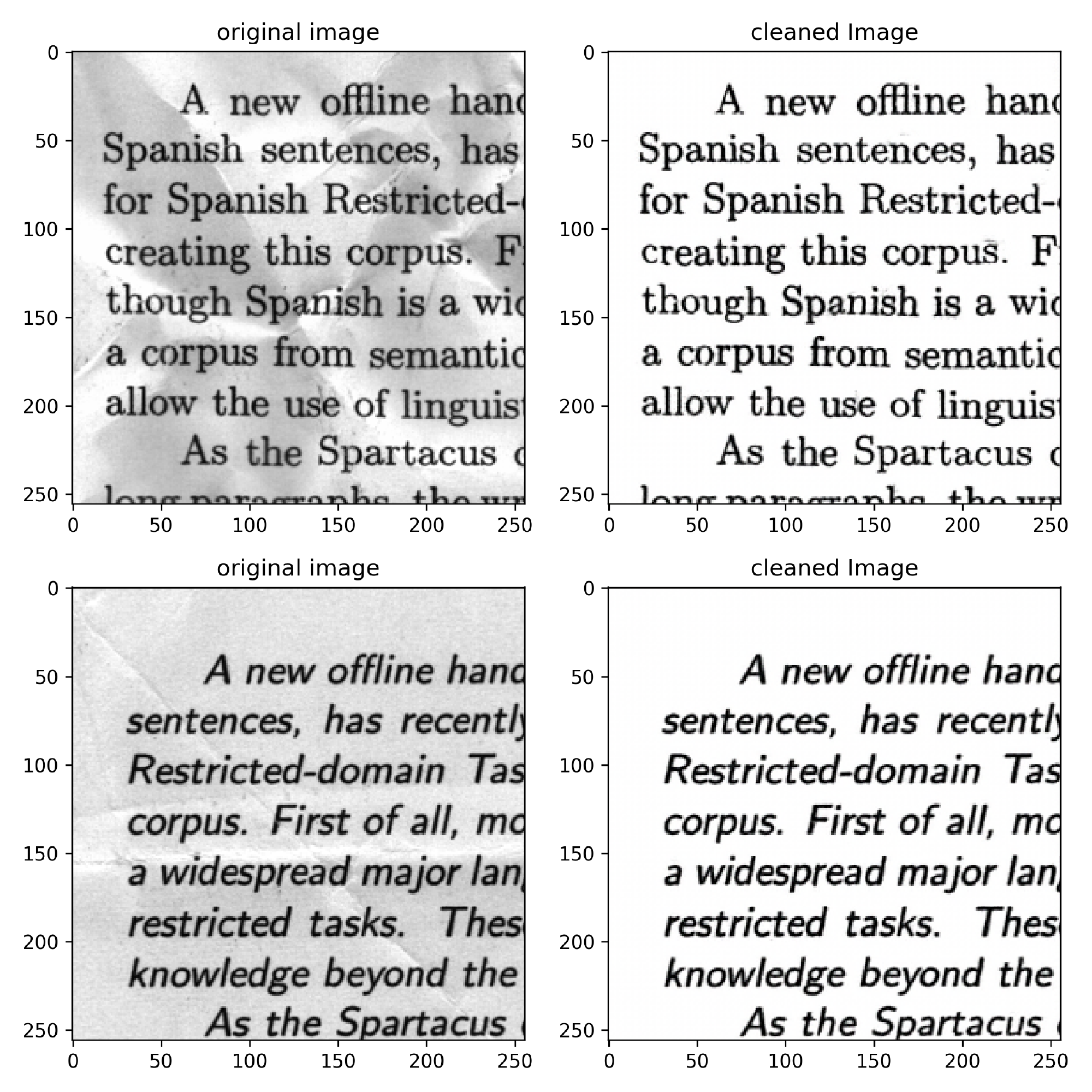}
\caption{Cleaned images of the \textit{scanned noisy documents} dataset after fine-tuning the model with a small amount of examples. The cleaned result can be seen on the right side whereas the original image is presented on the left side. Plotted are two random samples of the test set. Note that the artifacts are visibly recognizable, but do not affect the OCR result as much as the synthetic dataset. For that reason the AWER is not evaluated after fine-tuning.}
\label{fig:finetuning}
\end{figure} 

Fig. \ref{fig:finetuning} shows the results of the image cleaning task on the new dataset. As can be seen, the background artifacts can be eliminated completely, while the letters remain sharp. 
\newline

\subsection{Cross-Attention Skip Connections}
\label{res:cross-att}
Adding cross-attention skip connections to the network is one novelty of the proposed model. These do not only enable the possibility to bridge preprocessed extra information to the encoder, but the base variant of the model increases its performance by bridging higher-layer representations to the decoder, which already shows beneficial effects during training. The network shows a faster convergence regarding the validation loss. Using conventional skip connections in deeper layers often lead to very small activations in the bottleneck. This could not be observed with cross-attention skip connections.  
Another point that motivates the cross-attention module is that textual information can be easily incorporated and optimally processed due to the implemented attention mechanism. 
To show this, the model is fine-tuned on two completely different tasks during one training process: \begin{itemize}
    \item Cleaning: Clean the image from artifacts
    \item Inversion: Invert the image, without removing artifacts 
\end{itemize} 
A textual command is additionally being fed to the model to let it differentiate, which task to be fulfilled. The textual information is converted to an embedding using a Sentence-Transformer~\cite{10.18653/v1/d19-1410}, which is fine-tuned on the SNLI corpus~\cite{10.48550/arxiv.1508.05326}. The model is used, considering that differently worded commands should lead to similar results. For simplicity a list of 10 sentences differing in the formulation per task is used. Nevertheless this can easily be done with a greater variety of sentence-embeddings in form of questions or larger commands and texts including up to 128 sentences. The embedded information fused with the deepest cross-attention module. As we use the base model to in-cooperate the information during fine-tuning, the textual information simply replaces parts of this representation from the penultimate layer that have been bridged to the decoder during the training of the base model, see Fig. \ref{fig:commands}.
\begin{figure}[h]	
\centering
\includegraphics[width=0.47\textwidth]{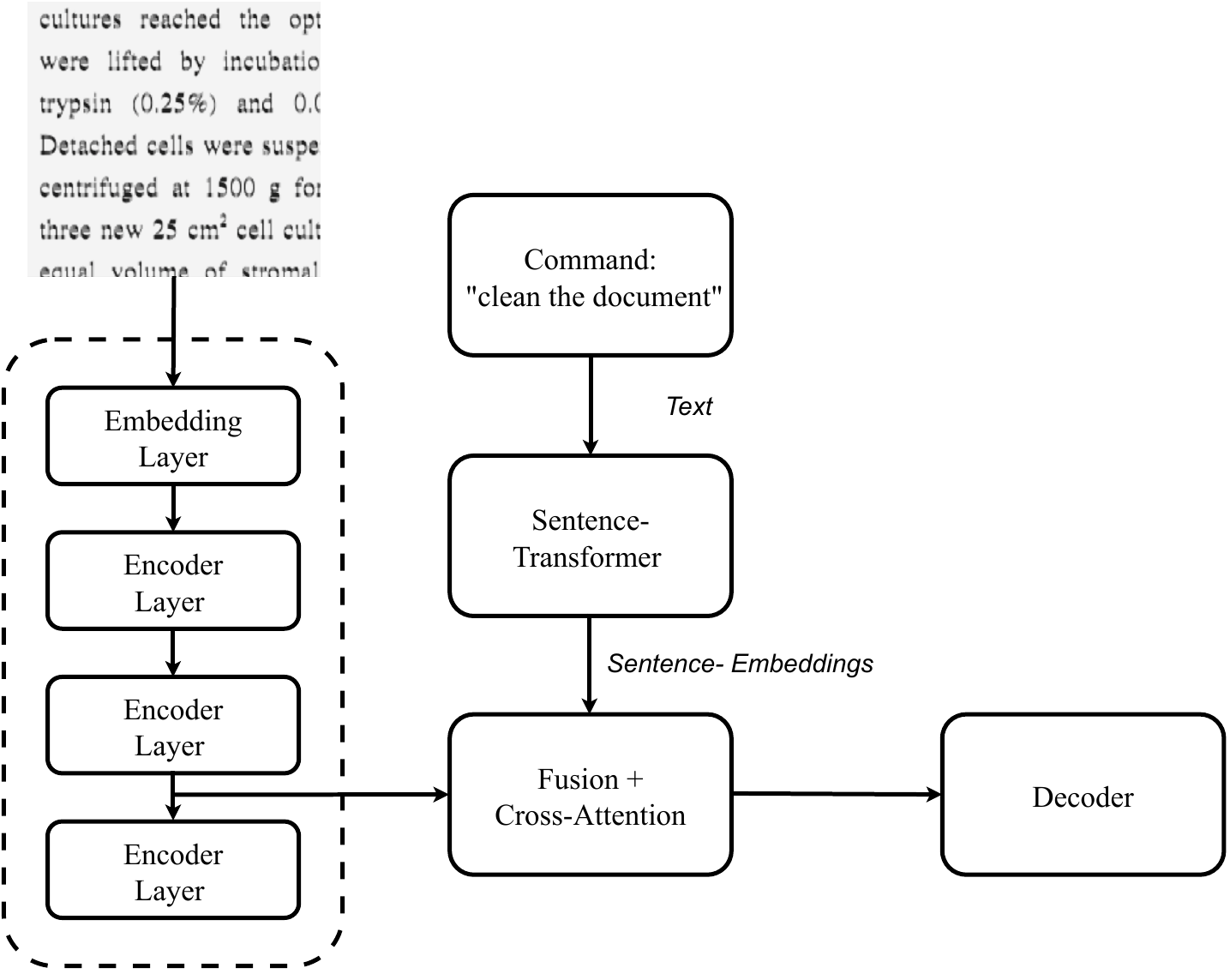}
\caption{Preparation of textual information and fusion in the cross-attention module. Commands are transformed to sentence-embeddings and added to the feature representation of the encoder. This can be done by replacing parts of the information to maintain dimensions or by adding the information, since one possesses flexibility in context dimensions. The first variant is used to simplify the fine-tuning process. }
\label{fig:commands}
\end{figure} 

Lost information may be a potential outcome of this approach. However, it should be noted that performing pre-training of the entire model from scratch using a similar task has the potential to yield more improved outcomes. However, the main purpose of the test is to determine whether the exploitation of cross-attention modules is suitable for recognizing the tasks by its textual description.  
The result shows that the model can be adapted in two different directions with a single training process using textual commands. 
A task recognition rate of $93.3\%$ testing on $30$ images when using random instructions of the set.

\section{Conclusion}
In this paper a Swin Transformer based UNet structure is presented. In order to establish a base model for artifact removal tasks, it undergoes pre-training using a novel synthetic dataset. The model shows good results on cleaning background noise and reconstructing pixel closures as well as character uniformity. The best performance is reached when dealing with noisy backgrounds bringing better OCR results, by enhancing the text quality on image level. Errors extracting the text can be reduced by up to 53.9 \%. 
It is shown that fine-tuning is, depending on the task, easy to execute without the need for large amounts of data or computing power. Especially due to the emergence of text-driven machine learning, the incorporation of modified skip connections has proven to be beneficial for enhancing the application flexibility.
It is shown that textual information can easily be integrated to the model, giving the opportunity to learn opposing tasks with only one model. 
Cross-attention skip connections not only bring the advantage of in-cooperating extra information (context) of any size to the decoder, but also show a faster convergence and more stability during training.
Like many transformer models based network structures one drawback of the proposed model is that it can become computationally expensive to train the model, leading to the necessity of smaller cropped images for investigating the model structure. Further strategies to maintain the performance with higher image sizes as well as a more specific pre-training to text data need to be considered in future investigations. Moreover, the synthetic dataset is planned to being extended and the loss function will be optimized to recognize the unreadable text in documents. 

\section*{Acknowledgments}
This work was supported by retensor GmbH and regrapes GmbH providing knowledge and hardware.

{\small
\bibliographystyle{ieee_fullname}
\bibliography{cited_papers}
}

\end{document}